# MULTI-VIEW RECURRENT NEURAL ACOUSTIC WORD EMBEDDINGS

**Wanjia He**
Department of Computer Science
University of Chicago
Chicago, IL 60637, USA
wanjia@ttic.edu

**Weiran Wang & Karen Livescu**
Toyota Technological Institute at Chicago
Chicago, IL 60637, USA
{weiranwang,klivescu}@ttic.edu

## ABSTRACT

Recent work has begun exploring neural acoustic word embeddings—fixed-dimensional vector representations of arbitrary-length speech segments corresponding to words. Such embeddings are applicable to speech retrieval and recognition tasks, where reasoning about whole words may make it possible to avoid ambiguous sub-word representations. The main idea is to map acoustic sequences to fixed-dimensional vectors such that examples of the same word are mapped to similar vectors, while different-word examples are mapped to very different vectors. In this work we take a multi-view approach to learning acoustic word embeddings, in which we jointly learn to embed acoustic sequences and their corresponding character sequences. We use deep bidirectional LSTM embedding models and multi-view contrastive losses. We study the effect of different loss variants, including fixed-margin and cost-sensitive losses. Our acoustic word embeddings improve over previous approaches for the task of word discrimination. We also present results on other tasks that are enabled by the multi-view approach, including cross-view word discrimination and word similarity.

## 1 INTRODUCTION

Word embeddings—continuous-valued vector representations of words—are an almost ubiquitous component of recent natural language processing (NLP) research. Word embeddings can be learned using spectral methods (Deerwester et al., 1990) or, more commonly in recent work, via neural networks (Bengio et al., 2003; Mnih & Hinton, 2007; Mikolov et al., 2013; Pennington et al., 2014). Word embeddings can also be composed to form embeddings of phrases, sentences, or documents (Socher et al., 2014; Kiros et al., 2015; Wieting et al., 2016; Iyyer et al., 2015).

In typical NLP applications, such embeddings are intended to represent the semantics of the corresponding words/sequences. In contrast, embeddings that represent the way a word or sequence *sounds* are rarely considered. In this work we address this problem, starting with embeddings of individual words. Such embeddings could be useful for tasks like spoken term detection (Fiscus et al., 2007), spoken query-by-example search (Anguera et al., 2014), or even speech recognition using a whole-word approach (Gemmeke et al., 2011; Bengio & Heigold, 2014). In tasks that involve comparing speech segments to each other, vector embeddings can allow more efficient and more accurate distance computation than sequence-based approaches such as dynamic time warping (Levin et al., 2013, 2015; Kamper et al., 2016; Settle & Livescu, 2016; Chung et al., 2016).

We consider the problem of learning vector representations of acoustic sequences and orthographic (character) sequences corresponding to single words, such that the learned embeddings represent the way the word sounds. We take a multi-view approach, where we jointly learn the embeddings for character and acoustic sequences. We consider several contrastive losses, based on learning from pairs of matched acoustic-orthographic examples and randomly drawn mismatched pairs. The losses correspond to different goals for learning such embeddings; for example, we might want the embeddings of two waveforms to be close when they correspond to the same word and far when they correspond to different ones, or we might want the distances between embeddings to correspond to some ground-truth orthographic edit distance.

One of the useful properties of this multi-view approach is that, unlike earlier work on acoustic word embeddings, it produces both acoustic and orthographic embeddings that can be directly compared. This makes it possible to use the same learned embeddings for multiple single-view and cross-view tasks. Our multi-view embeddings produce improved results over earlier work on acoustic word discrimination, as well as encouraging results on cross-view discrimination and word similarity.[1]

## 2 OUR APPROACH

In this section, we first introduce our approach for learning acoustic word embeddings in a multi-view setting, after briefly reviewing related approaches to put ours in context. We then discuss the particular neural network architecture we use, based on bidirectional long short-term memory (LSTM) networks (Hochreiter & Schmidhuber, 1997).

### 2.1 MULTI-VIEW LEARNING OF ACOUSTIC WORD EMBEDDINGS

Previous approaches have focused on learning acoustic word embeddings in a "single-view" setting. In the simplest approach, one uses supervision of the form "acoustic segment $\mathbf{x}$ is an instance of the word $\mathbf{y}$", and trains the embedding to be discriminative of the word identity. Formally, given a dataset of paired acoustic segments and word labels $\{(\mathbf{x}_i, \mathbf{y}_i)\}_{i=1}^N$, this approach solves the following optimization:

$$\min_{f,h} \ \mathrm{obj}_{classify} := \frac{1}{N} \sum_{i}^{N} \ell \left( h(f(\mathbf{x}_i)), \mathbf{y}_i \right), \tag{1}$$

where network $f$ maps an acoustic segment into a fixed-dimensional feature vector/embedding, $h$ is a classifier that predicts the corresponding word label from the label set of the training data, and the loss $\ell$ measures the discrepancy between the prediction and ground-truth word label (one can use any multi-class classification loss here, and a typical choice is the cross-entropy loss where $h$ has a softmax top layer). The two networks $f$ and $h$ are trained jointly. Equivalently, one could consider the composition $h(f(\mathbf{x}))$ as a classifier network, and use any intermediate layer's activations as the features. We refer to the objective in (1) as the "classifier network" objective, which has been used in several prior studies on acoustic word embeddings (Bengio & Heigold, 2014; Kamper et al., 2016; Settle & Livescu, 2016).

This objective, however, is not ideal for learning acoustic word embeddings. This is because the set of possible word labels is huge, and we may not have enough instances of each label to train a good classifier. In downstream tasks, we may encounter acoustic segments of words that did not appear in the embedding training set, and it is not clear that the classifier-based embeddings will have reasonable behavior on previously unseen words.

An alternative approach, based on Siamese networks (Bromley et al., 1993), uses supervision of the form "segment $\mathbf{x}^1$ is similar to segment $\mathbf{x}^2$, and is not similar to segment $\mathbf{x}^3$", where two segments are considered similar if they have the same word label and dissimilar otherwise. Models based on Siamese networks have been used for a variety of representation learning problems in NLP (Hu et al., 2014; Wieting et al., 2016), vision (Hadsell et al., 2006), and speech (Synnaeve et al., 2014; Kamper et al., 2015) including acoustic word embeddings (Kamper et al., 2016; Settle & Livescu, 2016). A typical objective in this category enforces that the distance between $(\mathbf{x}^1, \mathbf{x}^3)$ is larger than the distance between $(\mathbf{x}^1, \mathbf{x}^2)$ by some margin:

$$\min_{f} \ \mathrm{obj}_{siamese} := \frac{1}{N} \sum_{i}^{N} \max \left(0, \ m + dis\left(f(\mathbf{x}_i^1), f(\mathbf{x}_i^2)\right) - dis\left(f(\mathbf{x}_i^1), f(\mathbf{x}_i^3)\right)\right), \tag{2}$$

where the network $f$ extracts the fixed-dimensional embedding, the distance function $dis(\cdot, \cdot)$ measures the distance between the two embedding vectors, and $m > 0$ is the margin parameter. The term "Siamese" (Bromley et al., 1993; Chopra et al., 2005) refers to the fact that the triplet $(\mathbf{x}^1, \mathbf{x}^2, \mathbf{x}^3)$ share the same embedding network $f$.

Unlike the classification-based loss, the Siamese network loss does not enforce hard decisions on the label of each segment. Instead it tries to learn embeddings that respect distances between word pairs, which can be helpful for dealing with unseen words. The Siamese network approach also uses more examples in training, as one can easily generate many more triplets than (segment, label) pairs, and it is not limited to those labels that occur a sufficient number of times in the training set.

[1]Our tensorflow implementation is available at `https://github.com/opheadacheh/Multi-view-neural-acoustic-words-embeddings`

The above approaches treat the word labels as discrete classes, which ignores the similarity between different words, and does not take advantage of the more complex information contained in the character sequences corresponding to word labels. The orthography naturally reflects some aspects of similarity between the words' pronunciations, which should also be reflected in the acoustic embeddings. One way to learn features from multiple sources of complementary information is using a multi-view representation learning setting. We take this approach, and consider the acoustic segment and the character sequence to be two different views of the pronunciation of the word.

While many deep multi-view learning objectives are applicable (Ngiam et al., 2011; Srivastava & Salakhutdinov, 2014; Sohn et al., 2014; Wang et al., 2015), we consider the multi-view contrastive loss objective of (Hermann & Blunsom, 2014), which is simple to optimize and implement and performs well in practice. In this algorithm, we embed acoustic segments $\mathbf{x}$ by a network $f$ and character label sequences $\mathbf{c}$ by another network $g$ into a common space, and use weak supervision of the form "for paired segment $\mathbf{x}^+$ and its character label sequence $\mathbf{c}^+$, the distance between their embedding is much smaller than the distance between embeddings of $\mathbf{x}^+$ and an unmatched character label sequence $\mathbf{c}^-$". Formally, we optimize the following objective with such supervision:

$$\min_{f,g} \ \mathrm{obj}^0 := \frac{1}{N}\sum_{i}^{N} \max\left(0,\ m + dis\left(f(\mathbf{x}_i^+),\ g(\mathbf{c}_i^+)\right) - dis\left(f(\mathbf{x}_i^+),\ g(\mathbf{c}_i^-)\right)\right), \tag{3}$$

where $\mathbf{c}_i^-$ is a negative character label sequence of $\mathbf{x}_i^+$ to be contrasted with the positive/correct character sequence $\mathbf{c}_i^+$, and $m$ is the margin parameter. In this paper we use the cosine distance, $dis\left(\mathbf{a}, \mathbf{b}\right) = 1 - \left\langle \frac{\mathbf{a}}{\|\mathbf{a}\|},\ \frac{\mathbf{b}}{\|\mathbf{b}\|} \right\rangle$.[2]

Note that in the multi-view setting, we have multiple ways of generating triplets that contain one positive pair and one negative pair each. Below are the other three objectives we explore in this paper:

$$\min_{f,g} \ \mathrm{obj}^1 := \frac{1}{N}\sum_{i}^{N} \max\left(0,\ m + dis\left(f(\mathbf{x}_i^+),\ g(\mathbf{c}_i^+)\right) - dis\left(g(\mathbf{c}_i^+),\ g(\mathbf{c}_i^-)\right)\right), \tag{4}$$

$$\min_{f,g} \ \mathrm{obj}^2 := \frac{1}{N}\sum_{i}^{N} \max\left(0,\ m + dis\left(f(\mathbf{x}_i^+),\ g(\mathbf{c}_i^+)\right) - dis\left(f(\mathbf{x}_i^-),\ g(\mathbf{c}_i^+)\right)\right), \tag{5}$$

$$\min_{f,g} \ \mathrm{obj}^3 := \frac{1}{N}\sum_{i}^{N} \max\left(0,\ m + dis\left(f(\mathbf{x}_i^+),\ g(\mathbf{c}_i^+)\right) - dis\left(f(\mathbf{x}_i^+),\ f(\mathbf{x}_i^-)\right)\right). \tag{6}$$

$\mathbf{x}_i^-$ in (5) and (6) refers to a negative acoustic feature sequence, that is one with a different label from $\mathbf{x}_i^+$. We note that $\mathrm{obj}^1$ and $\mathrm{obj}^3$ contain distances between same-view embeddings, and are less thoroughly explored in the literature. We will also consider combinations of $\mathrm{obj}^0$ through $\mathrm{obj}^3$.

Finally, thus far we have considered losses that do not explicitly take into account the degree of difference between the positive and negative pairs (although the learned embeddings may implicitly learn this through the relationship between sequences in the two views). We also consider a cost-sensitive objective designed to explicitly arrange the embedding space such that word similarity is respected. In (3), instead of a fixed margin $m$, we use:

$$m(\mathbf{c}^+,\ \mathbf{c}^-) := m_{\max} \cdot \frac{\min\left(t_{\max},\ editdis(\mathbf{c}^+,\ \mathbf{c}^-)\right)}{t_{\max}}, \tag{7}$$

where $t_{\max} > 0$ is a threshold for edit distances (all edit distances above $t_{\max}$ are considered equally bad), and $m_{max}$ is the maximum margin we impose. The margin is set to $m_{max}$ if the edit distance between two character sequences is above $t_{\max}$; otherwise it scales linearly with the edit distance $editdis(\mathbf{c}^+,\ \mathbf{c}^-))$. We use the Levenshtein distance as the edit distance. Here we explore the cost-sensitive margin with $\mathrm{obj}^0$, but it could in principle be used with other objectives as well.

[2]In experiments, we use the unit-length vector $\frac{\mathbf{a}}{\|\mathbf{a}\|}$ as the embedding. It tends to perform better than $f(\mathbf{x})$ and more directly reflects the cosine similarity. This is equivalent to adding a nonlinear normalization layer on top of $f$.

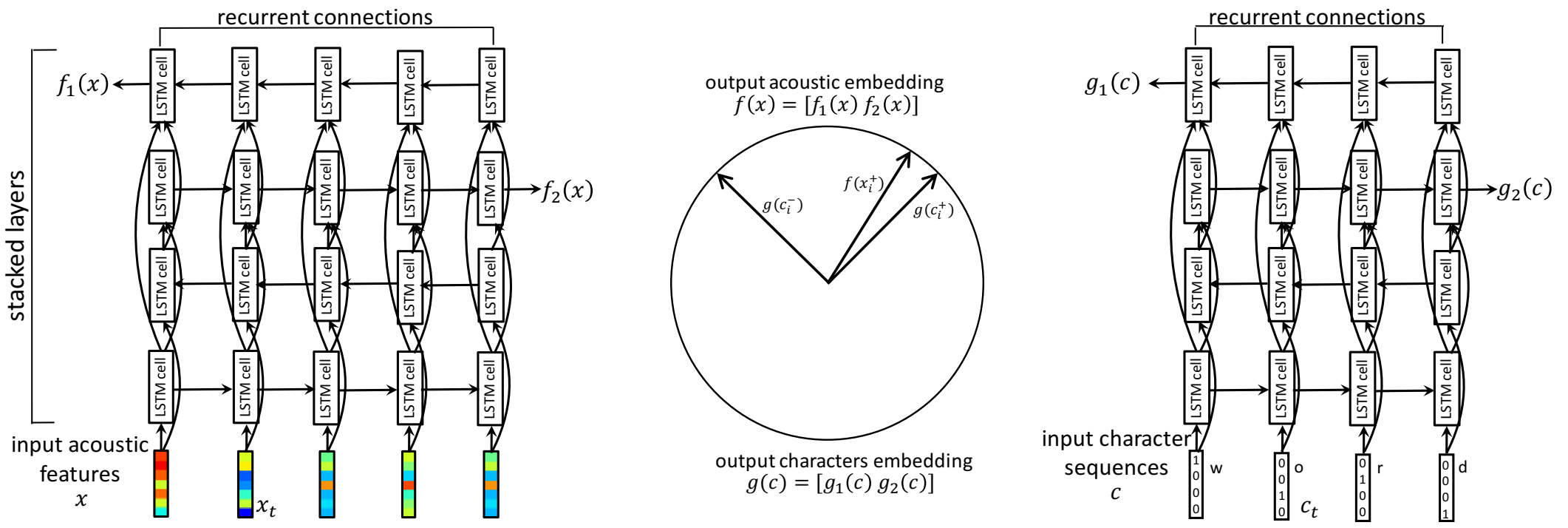

Figure 1: Illustration of our embedding architecture and contrastive multi-view approach.

### 2.2 Recurrent Neural Network Architecture

Since the inputs of both views have a sequential structure, we implement both $f$ and $g$ with recurrent neural networks and in particular long-short term memory networks (LSTMs). Recurrent neural networks are the state-of-the-art models for a number of speech tasks including speech recognition Graves et al. (2013), and LSTM-based acoustic word embeddings have produced the best results on one of the tasks in our experiments (Settle & Livescu, 2016).

As shown in Figure 1, our $f$ and $g$ are produced by multi-layer (stacked) bidirectional LSTMs. The inputs can be any frame-level acoustic feature representation and vector representation of the characters in the orthographic input. At each layer, two LSTM cells process the input sequence from left to right and from right to left respectively. At intermediate layers, the outputs of the two LSTMs at each time step are concatenated to form the input sequence to the next layer. At the top layer, the last time step outputs of the two LSTMs are concatenated to form a fixed-dimensional embedding of the view, and the embeddings are then used to calculate the cosine distances in our objectives.

## 3 Related Work

We are aware of no prior work on multi-view learning of acoustic and character-based word embeddings. However, acoustic word embeddings learned in other ways have recently begun to be studied. Levin et al. (2013) proposed an approach for embedding an arbitrary-length segment of speech as a fixed-dimensional vector, based on representing each word as a vector of dynamic time warping (DTW) distances to a set of template words. This approach produced improved performance on a word discrimination task compared to using raw DTW distances, and was later also applied successfully for a query-by-example task (Levin et al., 2015). One disadvantage of this approach is that, while DTW handles the issue of variable sequence lengths, it is computationally costly and involves a number of DTW parameters that are not learned.

Kamper et al. (2016) and Settle & Livescu (2016) later improved on Levin *et al.*'s word discrimination results using convolutional neural networks (CNNs) and recurrent neural networks (RNNs) trained with either a classification or contrastive loss. Bengio & Heigold (2014) trained convolutional neural network (CNN)-based acoustic word embeddings for rescoring the outputs of a speech recognizer, using a loss combining classification and ranking criteria. Maas et al. (2012) trained a CNN to predict a semantic word embedding from an acoustic segment, and used the resulting embeddings as features in a segmental word-level speech recognizer. Harwath and Glass Harwath & Glass (2015); Harwath et al. (2016); Harwath & Glass (2017) jointly trained CNN embeddings of images and spoken captions, and showed that word-like unit embeddings can be extracted from the speech model. CNNs require normalizing the duration of the input sequences, which has typically been done via padding. RNNs, on the other hand, are more flexible in dealing with very different-length sequences. Chen et al. (2015) used long short-term memory (LSTM) networks with a classification loss to embed acoustic words for a simple (single-query) query-by-example search task. Chung et al. (2016) learned acoustic word embeddings based on recurrent neural network (RNN) autoencoders, and found that they improve over DTW for a word discrimination task similar to that of Levin et al. (2013). Audhkhasi et al. (2017) learned autoencoders for acoustic and written words, as well as a model for comparing the two, and applied these to a keyword search task.

Evaluation of acoustic word embeddings in downstream tasks such as speech recognition and search can be costly, and can obscure details of embedding models and training approaches. Most evaluations have been based on word discrimination – the task of determining whether two speech segments correspond to the same word or not – which can be seen as a proxy for query-by-example search (Levin et al., 2013; Kamper et al., 2016; Settle & Livescu, 2016; Chung et al., 2016). One difference between word discrimination and search/recognition tasks is that in word discrimination the word boundaries are given. However, prior work has been able to apply results from word discrimination Levin et al. (2013) to improve a query-by-example system without known word boundaries Levin et al. (2015), by simply applying their embeddings to non-word segments as well.

The only prior work focused on vector embeddings of character sequences explicitly aimed at representing their acoustic similarity is that of Ghannay et al. (2016), who proposed evaluations based on nearest-neighbor retrieval, phonetic/orthographic similarity measures, and homophone disambiguation. We use related tasks here, as well as acoustic word discrimination for comparison with prior work on acoustic embeddings.

## 4 Experiments and results

The ultimate goal is to gain improvements in speech systems where word-level discrimination is needed, such as speech recognition and query-by-example search. However, in order to focus on the content of the embeddings themselves and to more quickly compare a variety of models, it is desirable to have surrogate tasks that serve as intrinsic measures of performance. Here we consider three forms of evaluation, all based on measuring whether cosine distances between learned embeddings correspond well to desired properties.

In the first task, **acoustic word discrimination**, we are given a pair of acoustic sequences and must decide whether they correspond to the same word or to different words. This task has been used in several prior papers on acoustic word embeddings Kamper et al. (2015, 2016); Chung et al. (2016); Settle & Livescu (2016) and is a proxy for query-by-example search. For each given spoken word pair, we calculate the cosine distance between their embeddings. If the cosine distance is below a threshold, we output "yes" (same word), otherwise we output "no" (different words). The performance measure is the average precision (AP), which is the area under the precision-recall curve generated by varying the threshold and has a maximum value of 1.

In our multi-view setup, we embed not only the acoustic words but also the character sequences. This allows us to use our embeddings also for tasks involving comparisons between written and spoken words. For example, the standard task of spoken term detection (Fiscus et al., 2007) involves searching for examples of a given text query in spoken documents. This task is identical to query-by-example except that the query is given as text. In order to explore the potential of multi-view embeddings for such tasks, we design another proxy task, **cross-view word discrimination**. Here we are given a pair of inputs, one a written word and one an acoustic word segment, and our task is to determine if the acoustic signal is an example of the written word. The evalution proceeds analogously to the acoustic word discrimination task: We output "yes" if the cosine distance between the embeddings of the written and spoken sequences are below some threshold, and measure performance as the average precision (AP) over all thresholds.

Finally, we also would like to obtain a more fine-grained measure of whether the learned embeddings capture our intuitive sense of similarity between words. Being able to capture word similarity may also be useful in building query or recognition systems that fail gracefully and produce human-like errors. For this purpose we measure the rank correlation between embedding distances and character edit distances. This is analogous to the evaluation of semantic word embeddings via the rank correlation between embedding distances and human similarity judgments (Finkelstein et al., 2001; Hill et al., 2015). In our case, however, we do not use human judgments since the ground-truth edit distances themselves provide a good measure. We refer to this as the **word similarity** task, and we apply this measure to both pairs of acoustic embeddings and pairs of character sequence embeddings. Similar measures have been proposed by Ghannay et al. (2016) to evaluate acoustic word embeddings, although they considered only near neighbors of each word whereas we consider the correlation across the full range of word pairs.

In the experiments described below, we first focus on the acoustic word discrimination task for purposes of initial exploration and hyperparameter search, and then largely fix the models for evaluation using the cross-view word discrimination and word similarity measures.

### 4.1 DATA

We use the same experimental setup and data as in Kamper et al. (2015, 2016); Settle & Livescu (2016). The task and setup were first developed by (Carlin et al., 2011). The data is drawn from the Switchboard English conversational speech corpus (Godfrey et al., 1992). The spoken word segments range in duration from 50 to 200 frames (0.5 - 2 seconds). The train/dev/test splits contain 9971/10966/11024 pairs of acoustic segments and character sequences, corresponding to 1687/3918/3390 unique words. In computing the AP for the dev or test set, all pairs in the set are used, yielding approximately 60 million word pairs.

The input to the embedding model in the acoustic view is a sequence of 39-dimensional vectors (one per frame) of standard mel frequency cepstral coefficients (MFCCs) and their first and second derivatives. The input to the character sequence embedding model is a sequence of 26-dimensional one-hot vectors indicating each character of the word's orthography.

### 4.2 MODEL DETAILS AND HYPERPARAMETER TUNING

We experiment with different neural network architectures for each view, varying the number of stacked LSTM layers, the number of hidden units for each layer, and the use of single- or bidirectional LSTM cells. A coarse grid search shows that 2-layer bidirectional LSTMs with 512 hidden units per direction per layer perform well on the acoustic word discrimination task, and we keep this structure fixed for subsequent experiments (see Appendix A for more details). We use the outputs of the top-layer LSTMs as the learned embedding for each view, which is 1024-dimensional if bidirectional LSTMs are used.

In training, we use dropout on the inputs of the acoustic view and between stacked layers for both views. The architecture is illustrated in Figure 1. For each training example, our contrastive losses require a corresponding negative example. We generate a negative character label sequence by uniformly sampling a word label from the training set that is different from the positive label. We perform a new negative label sampling at the beginning of each epoch. Similarly, negative acoustic feature sequences are uniformly sampled from all of the differently labeled acoustic feature sequences in the training set.

The network weights are initialized with values sampled uniformly from the range $[-0.05, 0.05]$. We use the Adam optimizer (Kingma & Ba, 2015) for updating the weights using mini-batches of 20 acoustic segments, with an initial learning rate tuned over $\{0.0001, 0.001\}$. Dropout is used at each layer, with the rate tuned over $\{0, 0.2, 0.4, 0.5\}$, in which $0.4$ usually outperformed others. The margin in our basic contrastive objectives 0-3 is tuned over $\{0.3, 0.4, 0.5, 0.6, 0.7\}$, out of which $0.4$ and $0.5$ typically yield best results. For $\mathrm{obj}^0$ with the cost-sensitive margin, we tune the maximum margin $m_{\max}$ over $\{0.5, 0.6, 0.7\}$ and the threshold $t_{\max}$ over $\{9, 11, 13\}$. We train each model for up to 1000 epochs. The model that gives the best AP on the development set is used for evaluation on the test set.

### 4.3 EFFECTS OF DIFFERENT OBJECTIVES

We presented four contrastive losses (3)–(6) and potential combinations in Section 2.1. We now explore the effects of these different objectives on the word discrimination tasks.

Table 1 shows the development set AP for acoustic and cross-view word discrimination achieved using the various objectives. We tuned the objectives for the acoustic discrimination task, and then used the corresponding converged models for the cross-view task. Of the simple contrastive objectives, $\mathrm{obj}^0$ and $\mathrm{obj}^2$ (which involve only cross-view distances) slightly outperform the other two on the acoustic word discrimination task. The best-performing objective is the "symmetrized" objective $\mathrm{obj}^0 + \mathrm{obj}^2$, which significantly outperforms all individual objectives (and the combination of the four). Finally, the cost-sensitive objective is very competitive as well, while falling slightly short of the best performance. We note that a similar objective to our $\mathrm{obj}^0 + \mathrm{obj}^2$ was used by Vendrov et al. (2016) for the task of caption-image retrieval, where the authors essentially use all non-paired examples from the other view in the minibatch as negative examples (instead of random sampling one negative example as we do) to be contrasted with one paired example.

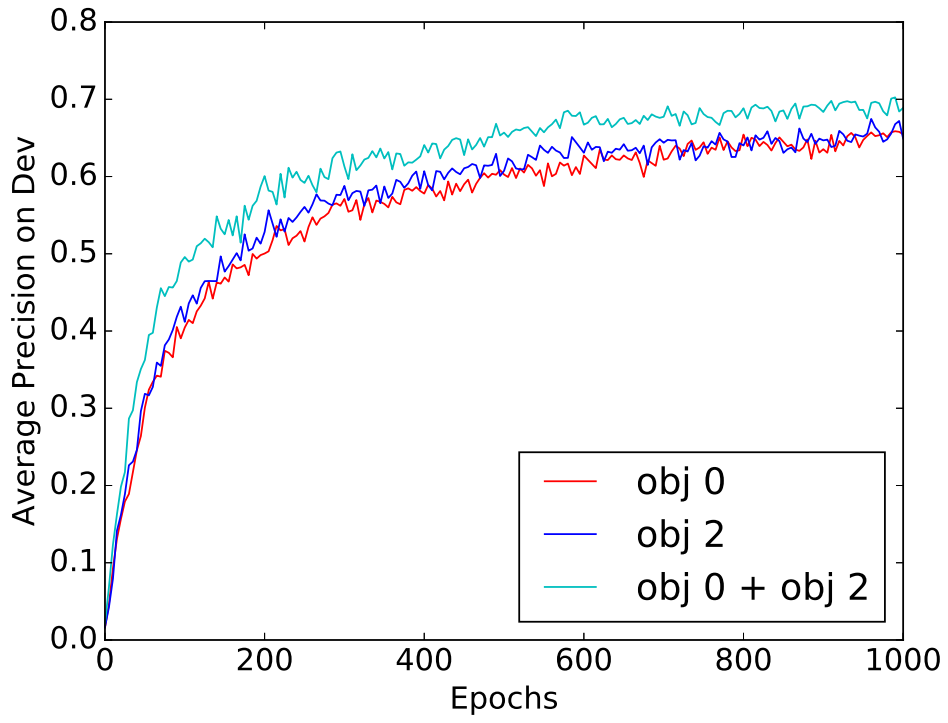

Figure 2: Development set AP for several objectives on acoustic word discrimination.

| **Objective** | **Dev AP** (acoustic) | **Dev AP** (cross-view) |
|---|---|---|
| $obj^0$ | 0.659 | 0.791 |
| $obj^1$ | 0.654 | 0.807 |
| $obj^2$ | 0.675 | 0.788 |
| $obj^3$ | 0.640 | 0.782 |
| $obj^0 + obj^2$ | 0.702 | 0.814 |
| $\sum_{i=0}^{3} obj^i$ | 0.672 | 0.804 |
| cost-sensitive | 0.671 | 0.802 |

Table 1: Word discrimination performance with different objectives.

| **Method** | **Test AP** (acoustic) | **Test AP** (cross-view) |
|---|---|---|
| MFCCs + DTW (Kamper et al., 2016) | 0.214 | |
| Correspondence autoencoder + DTW (Kamper et al., 2015) | 0.469 | |
| Phone posteriors + DTW (Carlin et al., 2011) | 0.497 | |
| Siamese CNN (Kamper et al., 2016) | 0.549 | |
| Siamese LSTM (Settle & Livescu, 2016) | 0.671 | |
| Our multi-view LSTM $\mathrm{obj}^0 + \mathrm{obj}^2$ | **0.806** | **0.892** |

Table 2: Final test set AP for different word discrimination approaches. The first line is a baseline using no word embeddings, but rather applying dynamic time warping (DTW) to the input MFCC features. The second and third lines are prior results using no word embeddings (but rather using DTW with learned correspondence autoencoder-based or phone posterior features, trained on larger external (in-domain) data). The remaining prior work corresponds to using cosine similarity between acoustic word embeddings.

Figure 2 shows the progression of the development set AP for acoustic word discrimination over $1000$ training epochs, using several of the objectives, where AP is evaluated every $5$ epochs. We observe that even after $1000$ epochs, the development set AP has not quite saturated, indicating that it may be possible to further improve performance.

Overall, our best-performing objective is the combined $\mathrm{obj}^0 + \mathrm{obj}^2$, and we use it for reporting final test-set results. Table 2 shows the test set AP for both the acoustic and cross-view tasks using our final model ("multi-view LSTM"). For comparison, we also include acoustic word discrimination results reported previously by Kamper et al. (2016); Settle & Livescu (2016). Previous approaches have not addressed the problem of learning embeddings jointly with the text view, so they can not be evaluated on the cross-view task.

### 4.4 Word similarity tasks

Table 3 gives our results on the word similarity tasks, that is the rank correlation (Spearman's $\rho$) between embedding distances and orthographic edit distance (Levenshtein distance between character sequences). We measure this correlation for both our acoustic word embeddings and for our text embeddings. In the case of the text embeddings, we could of course directly measure the Levenshtein distance between the inputs; here we are simply measuring how much of this information the text embeddings are able to retain.

| **Objective** | **$\rho$ (acoustic embedding)** | **$\rho$ (text embedding)** |
|---|---|---|
| fixed-margin ($\mathrm{obj}^0$) | 0.179 | 0.207 |
| cost-sensitive margin ($\mathrm{obj}^0$) | 0.240 | 0.270 |

Table 3: Word similarity results using fixed-margin and cost-sensitive objectives, given as rank correlation (Spearman's $\rho$) between embedding distances and orthographic edit distances.

Interestingly, while the cost-sensitive objective did not produce substantial gains on the word discrimination tasks above, it does greatly improve the performance on this word similarity measure. This is a satisfying observation, since the cost-sensitive loss is trying to improve precisely this relationship between distances in the embedding space and the orthographic edit distance.

Although we have trained our embeddings using orthographic labels, it is also interesting to consider how closely aligned the embeddings are with the corresponding phonetic pronunciations. For comparison, the rank correlation between our acoustic embeddings and phonetic edit distances is $0.226$, and for our text embeddings it is $0.241$, which are relatively close to the rank correlations with orthographic edit distance. A future direction is to directly train embeddings with phonetic sequence supervision rather than orthography; this setting involves somewhat stronger supervision, but it is easy to obtain in many cases.

Another interesting point is that the performance is not a great deal better for the text embeddings than for the acoustic embeddings, even though the text embeddings have at their disposal the text input itself. We believe this has to do with the distribution of words in our data: While the data includes a large variety of words, it does not include many very similar pairs. In fact, of all possible pairs of unique training set words, fewer than 2% have an edit distance below 5 characters. Therefore, there may not be sufficient information to learn to distinguish detailed differences among character sequences, and the cost-sensitive loss ultimately does not learn much more than to separate different words. In future work it would be interesting to experiment with data sets that have a larger variety of similar words.

### 4.5 Visualization of learned embeddings

Figure 3 gives a 2-dimensional t-SNE (van der Maaten & Hinton, 2008) visualization of selected acoustic and character sequences from the development set, including some that were seen in the training set and some previously unseen words. The previously seen words in this figure were selected uniformly at random among those that appear at least 15 times in the development set (the unseen words are the only six that appear at least 15 times in the development set). This visualization demonstrates that the acoustic embeddings cluster very tightly and are very close to the text embeddings, and that unseen words cluster nearly as well as previously seen ones.

While Figure 3 shows the relationship among the multiple acoustic embeddings and the text embeddings, the words are all very different so we cannot draw conclusions about the relationships between words. Figure 4 provides another visualization, this time exploring the relationship among the text embeddings of a number of closely related words, namely all development set words ending in "-ly", "-ing", and "-tion". This visualization confirms that related words are embedded close together, with the words sharing a suffix forming fairly well-defined clusters.

## 5 Conclusion

We have presented an approach for jointly learning acoustic word embeddings and their orthographic counterparts. This multi-view approach produces improved acoustic word embedding performance over previous approaches, and also has the benefit that the same embeddings can be applied for both spoken and written query tasks. We have explored a variety of contrastive objectives: ones with a fixed margin that aim to separate same and different word pairs, as well as a cost-sensitive loss that aims to capture orthographic edit distances. While the losses generally perform similarly for word discrimination tasks, the cost-sensitive loss improves the correlation between embedding distances and orthographic distances. One interesting direction for future work is to directly use knowledge about phonetic pronunciations, in both evaluation and training. Another direction is to extend our approach to directly train on both word and non-word segments.

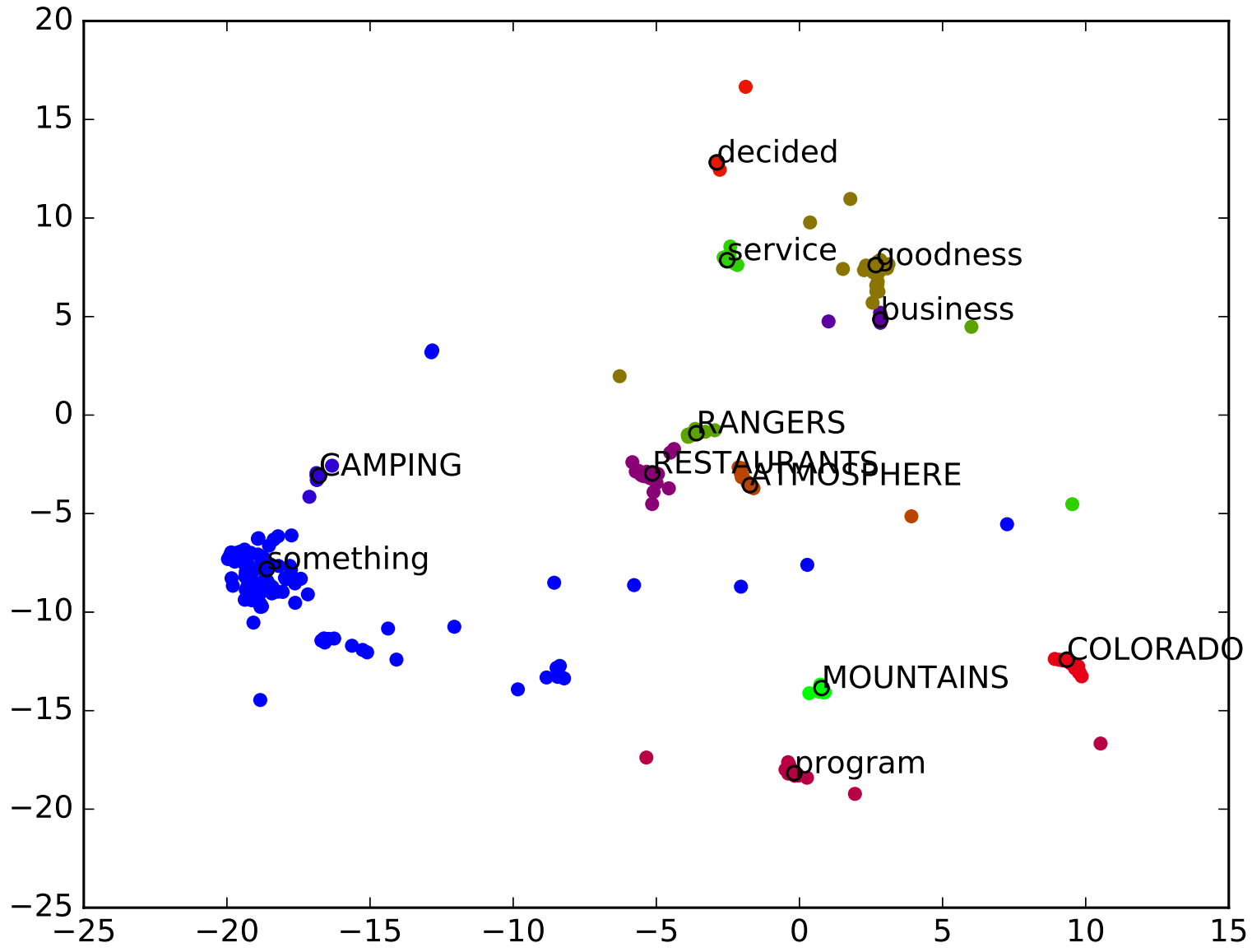

Figure 3: Visualization via t-SNE of acoustic word embeddings (colored markers) and corresponding character sequence embeddings (text), for a set of development set words with at least 15 acoustic tokens. Words seen in training are in lower-case; unseen words are in upper-case.

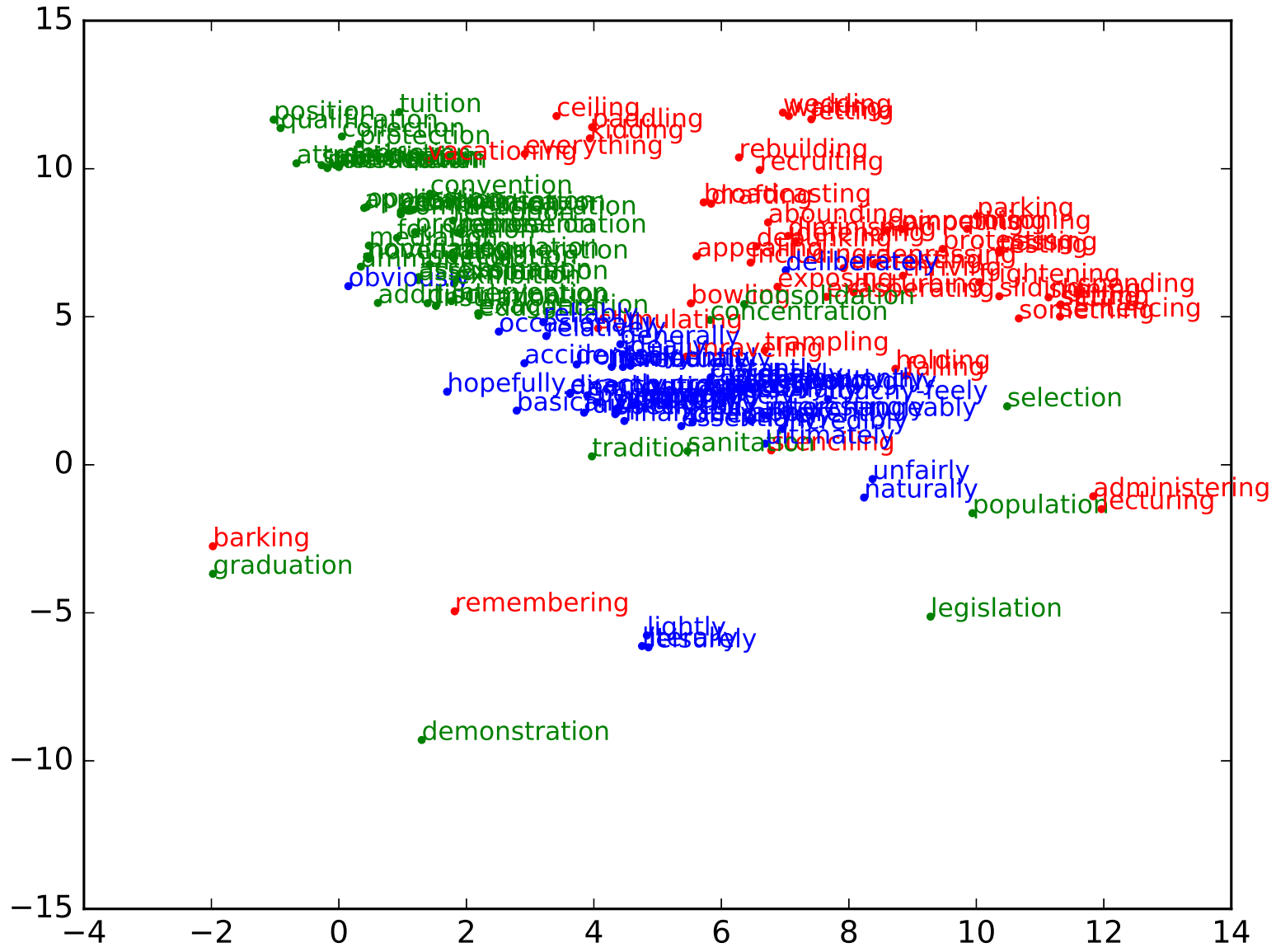

Figure 4: Visualization via t-SNE of character sequence embeddings for words with the suffixes “-ly” (blue), “-ing” (red), and “-tion” (green).

## Acknowledgments

This research was supported by a Google Faculty Award and by NSF grant IIS-1321015. The opinions expressed in this work are those of the authors and do not necessarily reflect the views of the funding agency. This research used GPUs donated by NVIDIA Corporation. We thank Herman Kamper and Shane Settle for their assistance with the data and experimental setup.

## REFERENCES

Xavier Anguera, Luis Javier Rodriguez-Fuentes, Igor Szöke, Andi Buzo, and Florian Metze. Query by example search on speech at mediaeval 2014. In *MediaEval*, 2014.

Kartik Audhkhasi, Andrew Rosenberg, Abhinav Sethy, Bhuvana Ramabhadran, and Brian Kingsbury. End-to-end ASR-free keyword search from speech. *arXiv preprint arXiv:1701.04313*, 2017.

Samy Bengio and Georg Heigold. Word embeddings for speech recognition. In *IEEE Int. Conf. Acoustics, Speech and Sig. Proc.*, 2014.

Yoshua Bengio, Réjean Ducharme, Pascal Vincent, and Christian Jauvin. A neural probabilistic language model. *Journal of Machine Learing Research*, 3(Feb):1137–1155, 2003.

Jane Bromley, Isabelle Guyon, Yann Lecun, Eduard Säckinger, and Roopak Shah. Signature verification using a siamese time delay neural network. In *Advances in Neural Information Processing Systems (NIPS)*, pp. 737–744, 1993.

Michael A Carlin, Samuel Thomas, Aren Jansen, and Hynek Hermansky. Rapid evaluation of speech representations for spoken term discovery. In *Proc. Interspeech*, 2011.

Guoguo Chen, Carolina Parada, and Tara N Sainath. Query-by-example keyword spotting using long short-term memory networks. In *Proc. ICASSP*, 2015.

Sumit Chopra, Raia Hadsell, and Yann LeCun. Learning a similarity metric discriminatively, with application to face verification. In *IEEE Computer Society Conf. Computer Vision and Pattern Recognition*, pp. 539–546, 2005.

Yu-An Chung, Chao-Chung Wu, Chia-Hao Shen, and Hung-Yi Lee. Unsupervised learning of audio segment representations using sequence-to-sequence recurrent neural networks. In *Proc. Interspeech*, 2016.

Scott Deerwester, Susan T Dumais, George W Furnas, Thomas K Landauer, and Richard Harshman. Indexing by latent semantic analysis. *Journal of the American society for information science*, 41 (6):391, 1990.

Lev Finkelstein, Evgeniy Gabrilovich, Yossi Matias, Ehud Rivlin, Zach Solan, Gadi Wolfman, and Eytan Ruppin. Placing search in context: The concept revisited. In *Proceedings of the 10th international conference on World Wide Web*, 2001.

Jonathan G Fiscus, Jerome Ajot, John S Garofolo, and George Doddingtion. Results of the 2006 spoken term detection evaluation. In *Proc. SIGIR*, volume 7, pp. 51–57. Citeseer, 2007.

Jort F Gemmeke, Tuomas Virtanen, and Antti Hurmalainen. Exemplar-based sparse representations for noise robust automatic speech recognition. *IEEE Transactions on Acoustics, Speech, and Language Processing*, 19(7):2067–2080, 2011.

Sahar Ghannay, Yannick Esteve, Nathalie Camelin, and Paul Deleglise. Evaluation of acoustic word embeddings. In *Proc. ACL Workshop on Evaluating Vector-Space Representations for NLP*, 2016.

John J Godfrey, Edward C Holliman, and Jane McDaniel. SWITCHBOARD: Telephone speech corpus for research and development. In *IEEE Int. Conf. Acoustics, Speech and Sig. Proc.*, 1992.

Alex Graves, Abdel rahman Mohamed, and Geoffrey Hinton. Speech recognition with deep recurrent neural networks. In *IEEE Int. Conf. Acoustics, Speech and Sig. Proc.*, 2013.

Raia Hadsell, Sumit Chopra, and Yann LeCun. Dimensionality reduction by learning an invariant mapping. In *IEEE Computer Society Conf. Computer Vision and Pattern Recognition*, 2006.

David Harwath and James Glass. Deep multimodal semantic embeddings for speech and images. In *Proc. IEEE Workshop on Automatic Speech Recognition and Understanding (ASRU)*, 2015.

David Harwath and James R Glass. Learning word-like units from joint audio-visual analysis. *arXiv preprint arXiv:1701.07481*, 2017.

David Harwath, Antonio Torralba, and James Glass. Unsupervised learning of spoken language with visual context. In *Advances in Neural Information Processing Systems (NIPS)*, 2016.

Karl Moritz Hermann and Phil Blunsom. Multilingual distributed representations without word alignment. In *Int. Conf. Learning Representations*, 2014. arXiv:1312.6173 [cs.CL].

Felix Hill, Roi Reichart, and Anna Korhonen. SimLex-999: Evaluating semantic models with (genuine) similarity estimation. *Computational Linguistics*, 41(4), 2015.

Sepp Hochreiter and Jürgen Schmidhuber. Long short-term memory. *Neural Computation*, 9(8): 1735–1780, 1997.

Baotian Hu, Zhengdong Lu, Hang Li, and Qingcai Chen. Convolutional neural network architectures for matching natural language sentences. In *Advances in Neural Information Processing Systems (NIPS)*, 2014.

Mohit Iyyer, Varun Manjunatha, Jordan Boyd-Graber, and Hal Daumé III. Deep unordered composition rivals syntactic methods for text classification. In *Proc. Association for Computational Linguistics*, 2015.

Herman Kamper, Micah Elsner, Aren Jansen, and Sharon J. Goldwater. Unsupervised neural network based feature extraction using weak top-down constraints. In *IEEE Int. Conf. Acoustics, Speech and Sig. Proc.*, 2015.

Herman Kamper, Weiran Wang, and Karen Livescu. Deep convolutional acoustic word embeddings using word-pair side information. In *IEEE Int. Conf. Acoustics, Speech and Sig. Proc.*, 2016.

Diederik Kingma and Jimmy Ba. ADAM: A method for stochastic optimization. In *Int. Conf. Learning Representations*, 2015.

Ryan Kiros, Yukun Zhu, Ruslan R Salakhutdinov, Richard Zemel, Raquel Urtasun, Antonio Torralba, and Sanja Fidler. Skip-thought vectors. In *Advances in Neural Information Processing Systems (NIPS)*, 2015.

Keith Levin, Katharine Henry, Aren Jansen, and Karen Livescu. Fixed-dimensional acoustic embeddings of variable-length segments in low-resource settings. In *Proc. IEEE Workshop on Automatic Speech Recognition and Understanding (ASRU)*, 2013.

Keith Levin, Aren Jansen, and Benjamin Van Durme. Segmental acoustic indexing for zero resource keyword search. In *IEEE Int. Conf. Acoustics, Speech and Sig. Proc.*, 2015.

Andrew L Maas, Stephen D Miller, Tyler M O'neil, Andrew Y Ng, and Patrick Nguyen. Word-level acoustic modeling with convolutional vector regression. In *Proc. ICML Workshop on Representation Learning*, 2012.

Tomas Mikolov, Ilya Sutskever, Kai Chen, Greg S Corrado, and Jeff Dean. Distributed representations of words and phrases and their compositionality. In *Advances in Neural Information Processing Systems (NIPS)*, 2013.

Andriy Mnih and Geoffrey Hinton. Three new graphical models for statistical language modelling. In *ICML*, 2007.

Jiquan Ngiam, Aditya Khosla, Mingyu Kim, Juhan Nam, Honglak Lee, and Andrew Ng. Multimodal deep learning. In *ICML*, pp. 689–696, 2011.

Jeffrey Pennington, Richard Socher, and Christopher D Manning. GloVe: Global vectors for word representation. In *Proc. Conference on Empirical Methods in Natural Language Processing*, 2014.

Shane Settle and Karen Livescu. Discriminative acoustic word embeddings: Recurrent neural network-based approaches. In *Proc. IEEE Workshop on Spoken Language Technology (SLT)*, 2016.

Richard Socher, Andrej Karpathy, Quoc V Le, Christopher D Manning, and Andrew Y Ng. Grounded compositional semantics for finding and describing images with sentences. *Transactions of the Association for Computational Linguistics*, 2:207–218, 2014.

Kihyuk Sohn, Wenling Shang, and Honglak Lee. Improved multimodal deep learning with variation of information. In *Advances in Neural Information Processing Systems (NIPS)*, pp. 2141–2149, 2014.

Nitish Srivastava and Ruslan Salakhutdinov. Multimodal learning with deep boltzmann machines. *Journal of Machine Learing Research*, pp. 2949–2980, 2014.

Gabriel Synnaeve, Thomas Schatz, and Emmanuel Dupoux. Phonetics embedding learning with side information. In *Proc. IEEE Workshop on Spoken Language Technology (SLT)*, 2014.

Laurens J. P. van der Maaten and Geoffrey E. Hinton. Visualizing data using $t$-SNE. *Journal of Machine Learing Research*, 9:2579–2605, November 2008.

Ivan Vendrov, Ryan Kiros, Sanja Fidler, and Raquel Urtasun. Order-embeddings of images and language. In *Int. Conf. Learning Representations*, 2016.

Weiran Wang, Raman Arora, Karen Livescu, and Jeff Bilmes. On deep multi-view representation learning. In *ICML*, pp. 1083–1092, 2015.

John Wieting, Mohit Bansal, Kevin Gimpel, and Karen Livescu. Towards universal paraphrastic sentence embeddings. In *Int. Conf. Learning Representations*, 2016.

# A Additional analysis

We first explore the effect of network architectures for our embedding models. We learn embeddings using objective $\mathrm{obj}^0$ and evaluate them on the acoustic and cross-view word discrimination tasks. The resulting average precisions on the development set are given in Table 4. All of the models were trained for 1000 epochs, except for the 1-layer unidirectional models which converged after 500 epochs. It is clear that bidirectional LSTMs are more successful than unidirectional LSTMs for these tasks, and two layers of LSTMs are much better than a single layer of LSTMs. We did not observe significant further improvement by using more than two layers of LSTMs. For all other experiments, we fix the architecture to 2-layer bidirectional LSTMs for each view.

| **Architecture** | **Dev AP** (acoustic word discrimination) | **Dev AP** (cross-view word discrimination) |
|---|---|---|
| 1-layer unidirectional | 0.379 | 0.616 |
| 1-layer bidirectional | 0.466 | 0.690 |
| 2-layer bidirectional | 0.659 | 0.791 |

Table 4: Average precision (AP) for acoustic and cross-view word discrimination tasks on the development set, using embeddings learned with objective $\mathrm{obj}^0$ and different LSTM architectures.

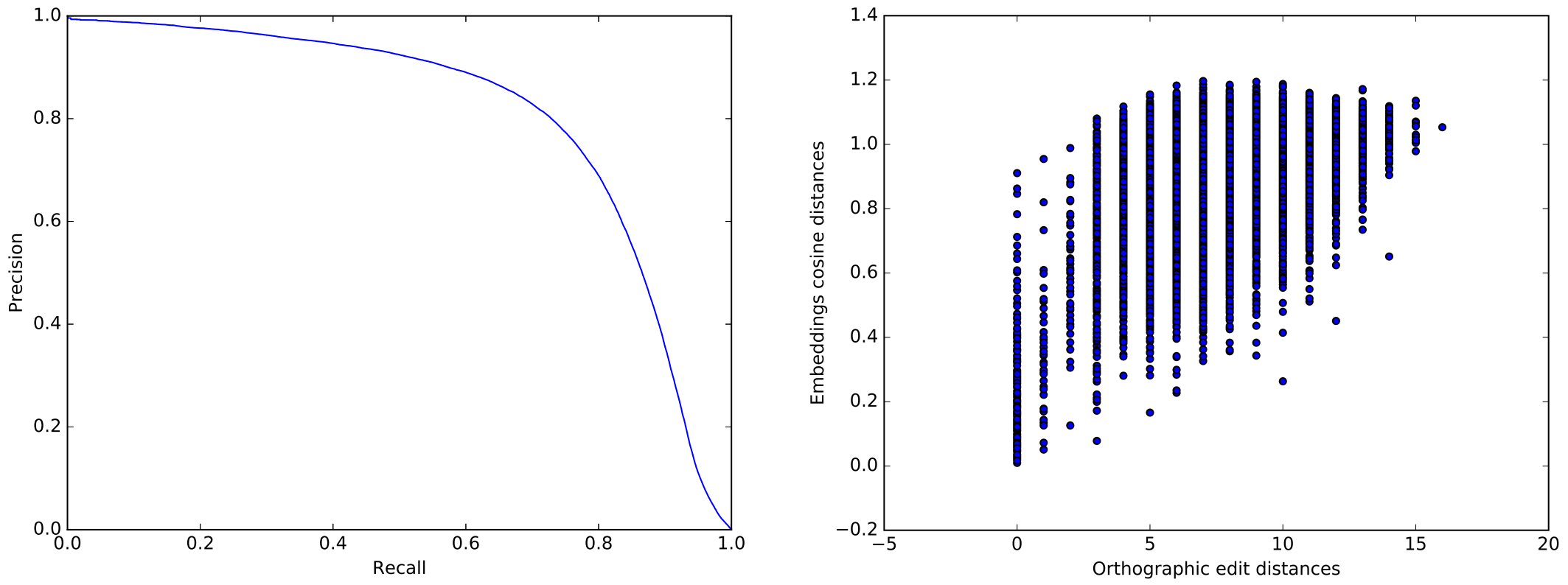

Figure 5: Precision-recall curve (left: two-layer bidirectional LSTM trained with $\mathrm{obj}^0 + \mathrm{obj}^2$ for word discrimination task) and scatter plot of embedding distances vs. orthographic distances (right: cost-sensitive margin model for word similarity task), for our best embedding models.

In Figure 5 we also give the precision-recall curve for our best models, as well as the scatter plot of cosine distances between acoustic embeddings vs. orthographic edit distances.